\pdfoutput=1

\documentclass[11pt]{article}

\usepackage{acl}

\usepackage{times}
\usepackage{latexsym}

\usepackage[T1]{fontenc}

\usepackage[utf8]{inputenc}

\usepackage{microtype}

\usepackage{inconsolata}

\usepackage{graphicx} 

\title{They Look Like Each Other: Case-based Reasoning for Explainable Depression Detection on Twitter using Large Language Models}

\author{Mohammad Saeid Mahdavinejad \\ Kansas State University  \\ \texttt{saeid@ksu.edu}
        \And  
        Peyman Adibi \\ University of Isfahan \\ \texttt{adibi@eng.ui.ac.ir}
        \AND  
        Amirhassan Monadjemi \\ National University of Singapore \\ \texttt{amir@comp.nus.edu.sg}
        \And  
        Pascal Hitzler \\ Kansas State University \\ \texttt{hitzler@ksu.edu}}

\usepackage{anyfontsize}

\begin{document}
\maketitle
\begin{abstract}
Depression is a common mental health issue that requires prompt diagnosis and treatment. Despite the promise of social media data for depression detection, the opacity of employed deep learning models hinders interpretability and raises bias concerns. We address this challenge by introducing ProtoDep, a novel, explainable framework for Twitter-based depression detection. ProtoDep leverages prototype learning and the generative power of Large Language Models to provide transparent explanations at three levels: (i) symptom-level explanations for each tweet and user, (ii) case-based explanations comparing the user to similar individuals, and (iii) transparent decision-making through classification weights. Evaluated on five benchmark datasets, ProtoDep achieves near state-of-the-art performance while learning meaningful prototypes. This multi-faceted approach offers significant potential to enhance the reliability and transparency of depression detection on social media, ultimately aiding mental health professionals in delivering more informed care.
\end{abstract}

\section{Introduction}
Depression is a common mental health disorder that affects a significant number of people worldwide. According to the National Institute of Mental Health, roughly 22.8\% of adults in the U.S. experience a diagnosable mental illness annually \citep{national-institute-of-mental-health-2023}. Timely diagnosis and intervention are crucial, as untreated or inadequately managed depression can lead to severe consequences, including suicide and chronic, risky behaviors like substance abuse \citep{GOODWIN2022726}.

Traditional methods for depression detection, heavily reliant on self-reported information through online questionnaires, are often hampered by low participation rates and potential selection bias. This has spurred the exploration of alternative approaches, with social media platforms emerging as a promising avenue \citep{chancellor2020methods, culotta2014estimating, de2014mental, guntuku2017detecting, paul2011you}.

While deep learning models exhibit significant potential in detecting depression on social media, their inherent "black box" nature presents a significant challenge \citep{ji2021mentalbert, nguyen2022improving}. This opacity hinders practitioners' ability to assess the validity of model predictions and raises concerns about potential biases or errors within the models themselves.

Recent efforts toward more explainable models for mental health assessment have primarily focused on two key approaches: post-hoc methods and interpretable models. Post-hoc methods aim to explain the predictions of pre-trained models retrospectively. However, they rely on approximations of a model's internal decision-making process, failing to explain why specific input features are crucial \citep{nguyen2021effectiveness}.

In contrast, interpretable models are inherently designed to be transparent by restricting their complexity. However, current state-of-the-art models primarily rely on attention weights for explanation \citep{han2022hierarchical}. The validity of these weights as reliable explanations remains under debate \citep{bibal2022attention}. Furthermore, these explanations often focus on low-level input features, like individual posts or tweets, which may not align with the higher-level concepts, such as symptoms, used by professionals.

This paper introduces ProtoDep, a novel explainable framework for depression detection on Twitter. It leverages prototype learning, utilizing representative data points (prototypes) to classify new instances. This framework facilitates the identification of key factors contributing to users' depressive behavior on social media through three distinct levels of explanations. Firstly, by harnessing the generative power of Large Language Models (LLMs), ProtoDep generates symptom-level explanations for each tweet and user, resembling human-readable concepts employed in mental health assessments. These explanations express the likelihood of each depression symptom being present for the specific tweet and user. Secondly, the framework provides a case-based explanation for the final prediction by comparing the user profile to the top k most representative users within the dataset. Finally, ProtoDep's classification weights reveal its transparent decision-making process.

Our evaluation on five benchmark datasets demonstrates that ProtoDep not only achieves near state-of-the-art performance compared to "black-box" models but also learns meaningful prototypes. The key contributions of this framework are as follows:
\begin{enumerate}
    \item Novel explainable framework for mental health assessment on social media.
    \item Multi-level explanations: (i) explaining underlying user symptoms, (ii) identifying similar users, and (iii) transparent decision-making.
    \item Maintains performance on five benchmark datasets.
    \item Leverages LLMs to learn meaningful prototypes.
\end{enumerate}

By addressing the critical gap in explainability through its multi-faceted approach, ProtoDep holds the potential to significantly improve the reliability and transparency of depression detection on social media, ultimately aiding mental health professionals in providing more informed care.

\section{Related Work}
The rise of social media has created exciting opportunities for mental health research. Its real-time nature and extensive archives offer unique advantages over traditional, retrospective studies. Researchers can track and potentially predict risk factors over time, enabling timely interventions for vulnerable communities \citep{kruzan2022social, livingston2014another, ridout2018use}.

This potential has driven exploration into using social media data to identify and predict mental health challenges like anxiety \citep{ahmed2022machine, saifullah2021comparison, shen2017detecting} and depression \citep{de2013social, de2013predicting, park2013perception, tsugawa2015recognizing, xu2021leveraging}.

Initial research employed basic methods for analyzing social media data to glean mental health insights \citep{coppersmith2014measuring, de2013social, coppersmith2014measuring}. Subsequently, the focus shifted towards developing feature engineering techniques and machine learning models for prediction \citep{birnbaum2017collaborative, moreno2011feeling, nguyen2014affective, rumshisky2016predicting, tsugawa2015recognizing}. For instance, \cite{de2013predicting} extracted linguistic features to build an SVM model for depression prediction.

The advent of deep learning further revolutionized the field, eliminating the need for hand-crafted features \citep{ji2018supervised, sawhney2018exploring}. \cite{tadesse2019detection} leveraged an LSTM-CNN model with word embeddings to identify suicide ideation on Reddit, showcasing the power of deep learning approaches.

Pre-trained language models (PLMs) have gained significant traction in natural language processing (NLP) tasks, including mental health prediction \citep{han2022hierarchical, ji2021mentalbert, nguyen2022improving}. \cite{jiang2020detection} utilized BERT's contextual representations for mental health issue detection, while \cite{otsuka2023diagnosing} evaluated BERT-based models in clinical settings. Additionally, multi-task learning approaches have been explored to predict multiple mental health conditions simultaneously \citep{benton2017multi}. \cite{sarkar2022predicting} trained a multi-task model for predicting both depression and anxiety, demonstrating the potential for joint prediction. However, these multi-task models often lack the flexibility to adapt to new tasks.

Despite these advancements, explainability remains a critical challenge in computational mental health prediction. While feature importance methods like SHAP \citep{shapley1953value, datta2016algorithmic, lundberg2017unified} and LIME \citep{ribeiro2016should} provide insights into feature contributions, they lack explanations for why specific features are crucial. Recent studies suggest that example-based explanations, often utilizing Case-Based Reasoning (CBR) techniques, resonate more with human users \citep{nguyen2021effectiveness}. However, this approach is limited by the underlying model architecture.

Building upon these developments, our work introduces a novel framework to depression detection on Twitter using prototype learning. Our key focus is to provide interpretable explanations at multiple levels, addressing the critical gap in explainability for mental health prediction on social media.

\section{Method}

\begin{figure*}[tb]
    \centering
    \includegraphics[width=1\textwidth]{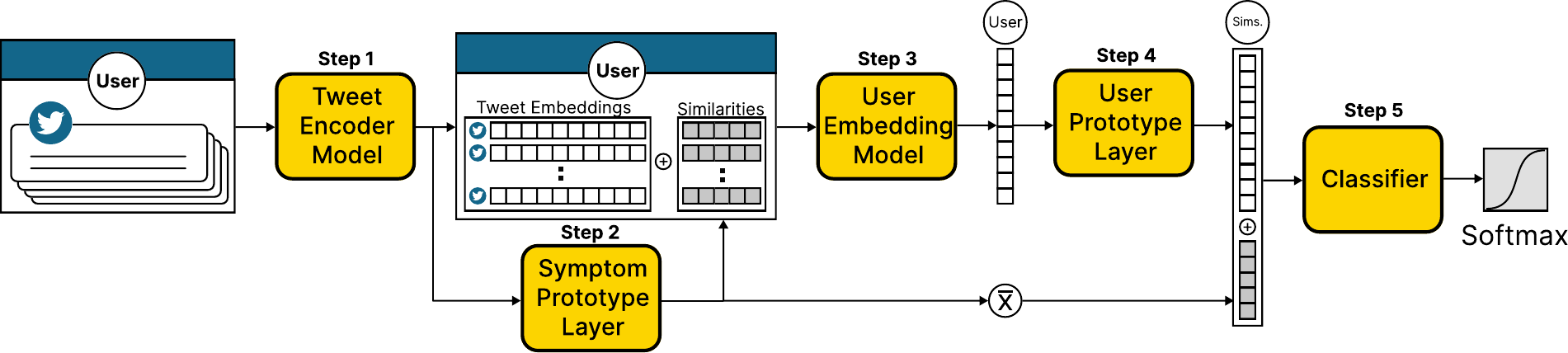}
    \caption{Overview of the Proto-Dep}
    \label{fig:modelDiagram}
\end{figure*}

This section presents ProtoDep, a novel framework for transparent reasoning about mental health in social media. ProtoDep uses prototypes to represent symptoms and users, enabling more interpretable explanations. Given a labeled user with its tweets, ProtoDep performs classification in five steps illustrated in Figure \ref{fig:modelDiagram}. \textbf{Step 1}: Embedding user tweets. \textbf{Step 2:} Learning symptom prototypes. \textbf{Step 3:} Encoding the user. \textbf{Step 4:} Learning user prototypes. \textbf{Step 5:} Performing classification.

\subsection{Preliminaries}

Our objective is to determine if a specific user $u$ is depressed. Each user has a collection of tweets $T$ where $T = \{t_1, t_2, ..., t_n\}$ and $n$ is number of tweets. Once the model is trained, the aim is to predict a binary label $\hat{y}$ for the user $u$, where $\hat{y} \in \{0, 1\}$. If $\hat{y} = 1$, the user is identified as depressed. Subsequently, $\hat{y}$ is matched against $y$, which is the ground truth.

\subsection{Step 1: Embedding User's Tweets}
Given a set of tweets $T$ for a user $u$, first, we obtain an embedding of $T$ using a pre-trained sentence encoder. $E$ will be an embedding matrix of $T$. Formally,  
\begin{equation}
E = \textnormal{TweetEncoder}(T)
\end{equation}

\subsection{Step 2: Learning Symptom Prototypes}
\begin{figure*}[tb]
    \centering
    \includegraphics[width=1\textwidth]{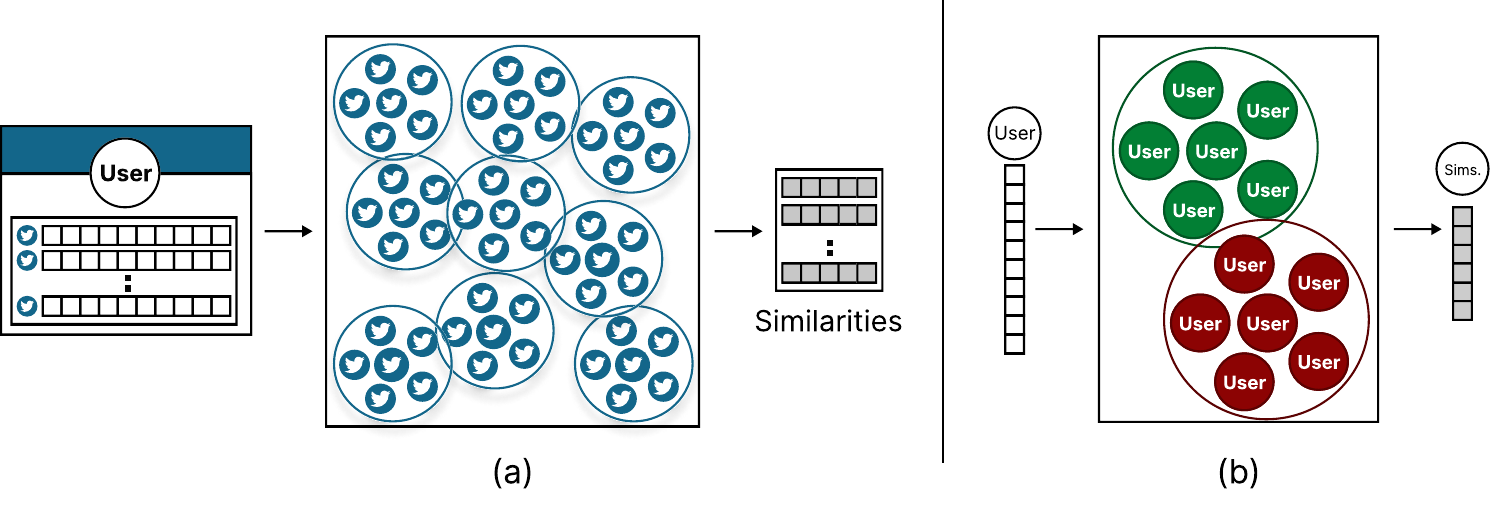}
    \caption{(a) Symptom Prototype Layer (b) User Prototype Layer}
    \label{fig:proto_layers}
\end{figure*}

This step focuses on training symptom prototypes that faithfully capture the essence of each depression symptom while maintaining close alignment with actual tweets from the dataset. In simpler terms, we aim to develop representations of symptoms that are both accurate and grounded in the language used by individuals describing their experiences. However, due to the absence of information about individual user symptoms or specific tweets (limited to user-level labels), we propose a supervised initialization strategy for prototypes, coupled with a specific loss function, to achieve this objective. This step encompasses two sub-steps: \textbf{A. Symptom Space Creation} and \textbf{B. Symptom Space Optimization}. Figure \ref{fig:proto_layers} $(a)$ illustrates a general schema for this layer.

\textbf{A. Symptom Space Creation:} 
The first step in creating an embedding space for symptom prototypes is identifying the underlying concepts or symptoms within the space. We use the Patient Health Questionnaire - 9 (PHQ-9\footnote{\url{https://www.apa.org/depression-guideline/patient-health-questionnaire.pdf}}), one of the most widely used questionnaires to assess depression \citep{KROENKE2010345}, as a reference for defining the concepts. The PHQ-9 is a self-administered questionnaire that measures the presence and severity of nine depressive symptoms over two weeks. It has been validated by multiple studies and is regarded as a reliable and accurate measure of depression \citep{Levisl1476}. The nine symptoms are Depressed Mood (S1), Loss of Interest or Pleasure (S2), Sleep Disturbance (S3), Fatigue or Low Energy (S4), Changes in Appetite (S5), Feelings of Guilt or Worthlessness (S6), Difficulty Concentrating (S7), Psychomotor Agitation or Retardation (S8), and Suicidal Thoughts (S9). We consider these symptoms as the base concepts in our embedding space to simulate human reasoning processes and enhance ProtoDep interpretability.

The second step is to initialize a set of prototypes for each concept. Manually creating exemplary sets for each prototype proves resource-intensive and needs iterative refinement. To address this challenge, we leverage the generative capabilities of LLMs. Specifically, we employ GPT-4 to automatically generate relevant examples, focusing on different aspects of a given symptom. These examples serve as our initial set of prototypes for subsequent training. We note that the number of prototypes is an important hyperparameter, and generating different numbers of examples from GPT-4 for each experiment is inconsistent and impractical. Therefore, we generate a maximum number of examples once and use the mean of the embedded examples as a base prototype for each symptom. Then, for each experiment, we sample around each base prototype with a normal distribution. We define $p_{base}^{j}$ as the base prototype and $P^{j}$ to be set of $m$ prototypes for the symptom $j$ by:
\begin{equation}
   P^{j} \sim \mathcal{N}(p_{base}^{j},\,\sigma^{2})\,.
\end{equation}
where $\sigma^{2}$ is variance. Therefore, $P$ will be set for all symptom prototypes. 

\textbf{B. Symptom Space Optimization:} 
Given the lack of tweet-level labels, we propose a novel approach that leverages supervised-initialized prototypes. Specifically, we formulate the optimization process as a multi-label classification task, where each tweet is labeled with the nearest symptom within the embedding space. By adopting this strategy, we effectively use prior knowledge from the initial symptom space while accommodating the lack of labeled tweets. As a result, we define the total symptom loss as the sum of two terms:
\begin{equation}
    L_{symp} = \lambda_1*L_{sinkhorn} + \lambda_2*L_{mse}
\end{equation}
where $\lambda_1$ and  $\lambda_2$ are hyperparameters, and $L_{sinkhorn}$ is Sinkhorn loss, a mathematical tool that computes optimal transport between two probability distributions \citep{cuturi2013sinkhorn}. This choice has several advantages over conventional loss functions. It enhances the stability and robustness of the training process, mitigates the impact of noise and overfitting, and accelerates the convergence rate \citep{feydy2019interpolating}. For calculating $L_{sinkhorn}$, first, we calculate a cosine similarity between a tweet embedding $e_i$ and a symptom $p_{k}^{j}$: 
\begin{equation}
    \textnormal{sim}_{i,j,k} = \cos(e_i,p_k^j)
\end{equation}
where $i\in\{1, ..., n\}$ and $j\in\{1, ..., 9\}$ and $k\in\{1, ..., m\}$. Then, we assign a label denoted as $s_i$ to each tweet based on its nearest symptom.
\begin{equation}
    s_{i} = \arg \max_{j} \frac{1}{m} \sum_{k=1}^{m} \textnormal{sim}_{i,j,k}
\end{equation} 
By defining all tweet embeddings with the same symptom as:
\begin{equation}
    E^j = \{e_i| s_i = j\}
\end{equation}
The $L_{sinkhorn}$ will be:
\begin{equation}
    L_{sinkhorn} = \frac{1}{9} \sum_{j=1}^{9} \textnormal{sinkhorn}(E^j,P^j)
\end{equation} 

Finally, $L_{mse}$ is a mean squared error loss and a measure of the difference between the input samples and their reconstructions using the nearest prototype. It encourages the prototypes to be representative of the input data. For this purpose, we find the index of the nearest prototype to each tweet embedding $c_i$ as:
\begin{equation}
    c_{i} = \arg \max_{j,k} \textnormal{sims}_{i,j,k}
\end{equation}
Next, we define the nearest prototype to $e_i$ as:
\begin{equation}
    p_{c_{i}} = \{p_k^j| (j,k) = c_i\}
\end{equation}
Now the $L_{mse}$ will be calculated as:
\begin{equation}
    L_{mse} = \frac{1}{n} \sum_{i=1}^{n} (e_i-p_{c_{i}})^2
\end{equation} 

\subsection{Step 3: Encoding the User}

Following the approach of \citep{han2022hierarchical}, we use a multi-layer attention mechanism and a feed-forward neural network to encode the sequential user behavior. The encoder model for the ProtoDep framework can vary depending on the problem domain and the data modality, which we will elaborate on in section~\ref{ablation}. We also append the similarity scores between the tweets and the symptom prototypes to improve the tweet representations. Our experiments show this can help the model learn better user representations and perform more accurately. Formally:
\begin{equation}
    \textnormal{SympSims} = \frac{1}{m} \sum_{k=1}^{m} \textnormal{sim}_{i,j,k}
\end{equation} 
And user embedding $e_{u}$ will be:
\begin{equation}
    e_{u} = \textnormal{UserEncoder}(E \oplus \textnormal{SympSims})
\end{equation} 

\subsection{Step 4: Learning User Prototypes}

This step provides transparent, case-based reasoning to evaluate the user’s depressive behavior. It follows the same principle as the learning symptom prototype step and consists of two sub-steps: \textbf{A. User Space Creation} and \textbf{B. User Space Optimization}. Figure \ref{fig:proto_layers} $(b)$ illustrates this step.

\textbf{A. User Space Creation:} 
Social media datasets for depression detection often exhibit an imbalance between the number of users or tweets in each class. This may negatively impact the reasoning of deep learning models as they may prioritize the majority class during training. Inspired by \citep{das-etal-2022-prototex}, we encourage the model to find the best examples for both classes to find a more effective decision boundary between them. Unlike the symptom prototype space, which relies on predefined prototypes, the user prototype space allows the model to learn the prototypes from the data. Consequently, we randomly initialize $k$ different vectors per class as initial prototypes.

\textbf{B. User Space Optimization:} 
In this step, we adopt the same optimization strategy as in step 2, but with a crucial difference. We leverage the user-level labels to learn the prototypes in a supervised fashion—this way, we do not require the computation of $s_i$. We denote the total loss for this step as $L_{user}$.

\subsection{Step 5: Classification}
The final step of our model is to classify the users based on their similarity to symptoms and user prototypes. First, we calculate the average of all tweet-symptom similarities, providing an overall measure of the similarity between a user's tweets and the symptom prototypes. Then, we concatenate these scores with the user prototype similarities and feed this into a linear layer followed by a Softmax function to obtain the final classification. We use binary cross-entropy (BCE) loss for this step, and the total loss function for our model will be:
\begin{equation}
    L = L_{symp} + L_{user} + L_{BCE}
\end{equation}

\section{Experiment Results}

\textbf{Dataset.} We employ an openly available Twitter dataset, MDL, specifically designed for depression detection. For comparison purposes, we use the specific version of the MDL dataset provided by \citep{han2022hierarchical}. In this dataset, individuals who posted tweets containing predefined phrases indicative of depression, such as "I'm," "I was," "I am," or "I've been diagnosed with depression," were labeled as depressive. Conversely, those who never posted any tweet containing the term "depress" were labeled as non-depressive. \citep{han2022hierarchical} employed a random selection of users to create five distinct datasets. The train, validation, and test sets utilize 60\%, 20\%, and 20\% of the entire dataset. This resulted in 2,524 users in the train set and 842 users in each validation and test set. More details regarding the dataset can be found in \citep{han2022hierarchical}.

\textbf{Baselines.} To evaluate the ProtoDep framework, we compare its performance to four established depression detection baselines. Given its similarity in approach, we consider \citep{han2022hierarchical} the most relevant and significant baseline for our model. Additionally, we compare our results to \citep{gui2019cooperative, lin2020sensemood, zhang2021mam}, as these studies appeared to be most pertinent to our work. 

\textbf{Setup.}  We trained all models using a GeForce RTX 3090 with 64GB of RAM and Pytorch 2.0.1. We tuned the hyperparameters on the validation data and optimized all neural models with the AdamW algorithm. The learning rate and the batch size were 1e-3 and 64, respectively. We applied early stopping based on the F1 score on the validation data. The maximum number of tweets per input was 200. For symptom and user prototypes, we set the number of prototypes per class to $m = 7$ and $k = 3$, respectively. We sampled normally around the base prototypes with $\sigma=0.025$. We also used two-layer attention for Step 3. We chose "all-mpnet-base-v2" \citep{reimers-2019-sentence-bert} as our Tweet Encoder Model. We discuss more on these in section \ref{ablation}.

\textbf{Result (1): Classification Performance.}
We assess ProtoDep using five benchmark datasets and compare it against state-of-the-art methods, as demonstrated in Table \ref{tab:performance}. ProtoDep achieves a competitive 94.4\% average F1 score, providing more intuitive and interpretable prototypes for classification decisions and maintaining consistent performance across different randomly sampled datasets (D1-D5). We also explore another variant of ProtoDep, ProtoDep-Acc, to demonstrate its power to achieve state-of-the-art performance. ProtoDep-Acc leverages a different loss function for learning prototypes. We discuss more about ProtoDep-Acc in section \ref{ablation}.

\begin{table}[]
\begin{tabular}{l|lll}
\hline
\textbf{Model} & \textbf{P} & \textbf{R} & \textbf{F1}  \\ \hline
\citep{gui2019cooperative}            & 0.900      & 0.901      & 0.900      \\
\citep{lin2020sensemood}            & 0.903      & 0.870      & 0.886          \\
\citep{zhang2021mam}          & 0.909      & 0.904      & 0.912           \\
\citep{han2022hierarchical}            & 0.975       & 0.969      & 0.972   \\      \hline
ProtoDep-Acc          & \textbf{0.985}          & \textbf{0.995}          & \textbf{0.990}          \\
\hline
ProtoDep (avg)       & 0.934      & 0.954      & 0.944                \\ \hline
D1             & 0.964      & 0.953      & 0.959                \\
D2             & 0.898      & 0.951      & 0.924                \\
D3             & 0.984      & 0.991      & 0.987                \\
D4             & 0.931      & 0.931      & 0.931                \\
D5             & 0.893      & 0.946      & 0.919                \\ \hline
\end{tabular}
\caption{Depression Detection Results. ProtoDep-Acc and ProtoDep (avg) are the average values across D1-D5.}
\label{tab:performance}
\end{table}

\textbf{Result (2): Explainable Prototypes.}
We evaluate the quality of ProtoDep's learned prototypes using two distinct methods. First, we compare these prototypes with manually labeled ground truth prototypes created by domain experts. To achieve this, we leverage a specialized dictionary developed by \citep{yazdavar2017semi}. This lexicon contains an extensive collection of depression-related terms specifically associated with the nine symptom categories of the PHQ-9. These terms have undergone meticulous curation to capture the subtle nuances inherent in depression symptoms.

\begin{figure}
    \centering
    \includegraphics[width=1\linewidth]{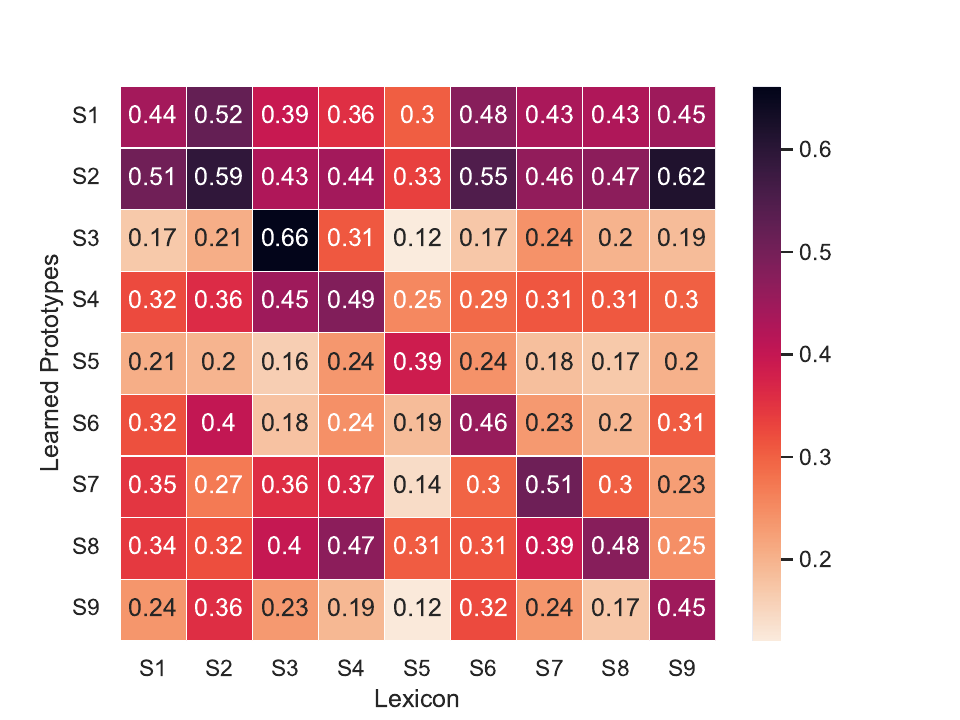}
    \caption{Similarity between learned symptom prototypes and ground truth lexicon.}
    \label{fig:lexicon_sims}
\end{figure}

By aligning our learned prototypes with this established lexicon, we can assess their relevance and meaningfulness within the context of depression. To quantify this alignment, we compute the mean representation for each symptom prototype and subsequently measure its cosine similarity with the embedded ground truth lexicon. Figure \ref{fig:lexicon_sims} visually illustrates this similarity, revealing a robust alignment between the learned symptom prototypes and the clinically relevant terms associated with depression. Notably, we focus on the diagonal elements of the similarity matrix, which correspond to the self-similarity scores. These high values indicate that our generated prototypes align with the clinically relevant terminology. Another observation drawn from the figure indicates a heightened similarity between the first two symptoms, specifically ‘Lack of Interest’ and ‘Feeling Down,’ and other symptoms. One plausible explanation for this phenomenon is the initialization bias as the initial GPT symptoms exhibited a similar pattern.

\begin{figure}
    \centering
    \includegraphics[width=1\linewidth]{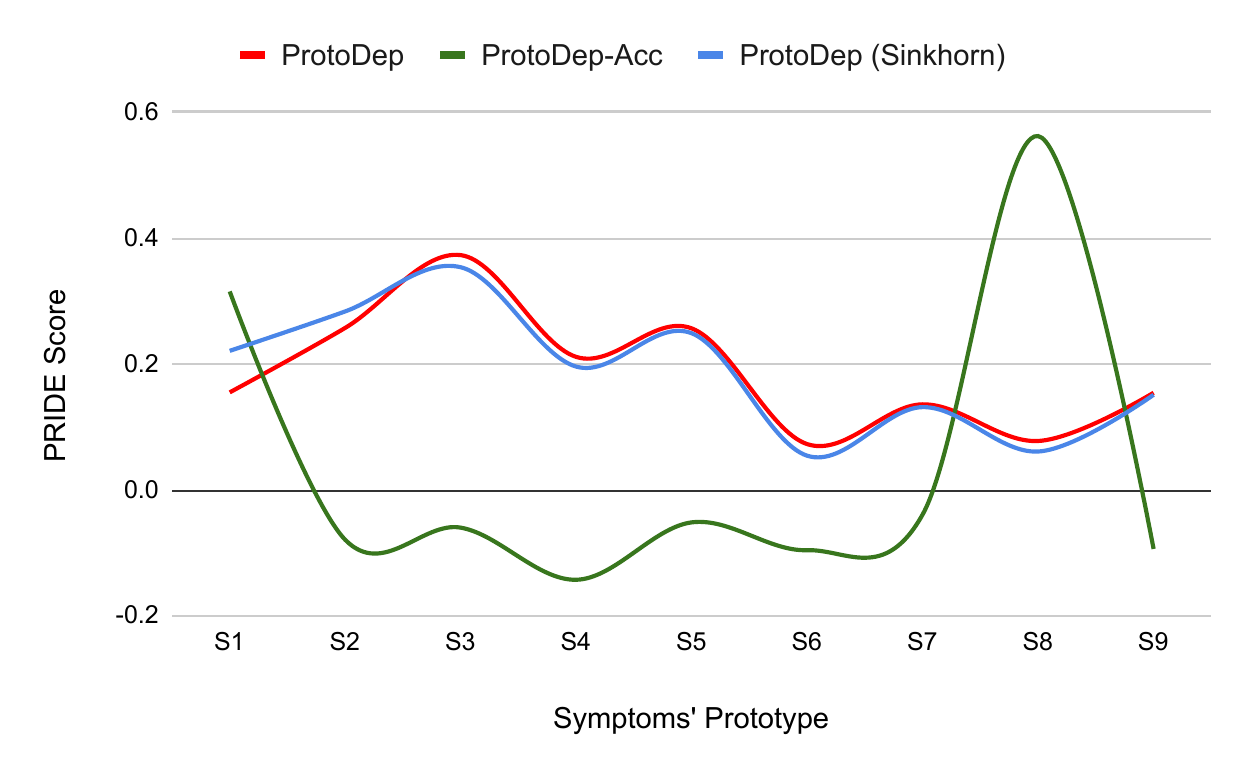}
    \caption{Visualization of the PRIDE score for learned prototypes for ProtoDep, ProtoDep-Acc, and ProtoDep (Sinkhorn) Models.}
    \label{fig:pride}
\end{figure}

Second, we assessed the prototypes' discriminative power by using the PRIDE score \citep{ni2022multimodal}, inspired by \citep{zhang2021prototype}. This method defines a "real" prototype for each category by averaging its data points and then measures the similarity between these real prototypes and the learned ones. A high PRIDE score indicates a learned prototype's effectiveness in capturing its designated category while differentiating itself from others. We assume the nearest tweet to each symptom in the ground truth lexicon is the real prototype in the dataset. Figure \ref{fig:pride} demonstrates that ProtoDep achieves positive PRIDE scores for all symptoms’ prototypes, implying effective learning of distinct and representative symptom prototypes.

We assess the efficacy of learned user prototypes by reporting the PRIDE score. Specifically, we identify a representative ‘real’ user for each class by computing the mean of all users within that class and subsequently calculating the PRIDE score. Notably, both depressed and non-depressed classes exhibit positive PRIDE scores (0.27 and 0.33, respectively), affirming that ProtoDep effectively captures a meaningful prototype space for users.

\begin{figure}
    \centering
    \includegraphics[width=1\linewidth]{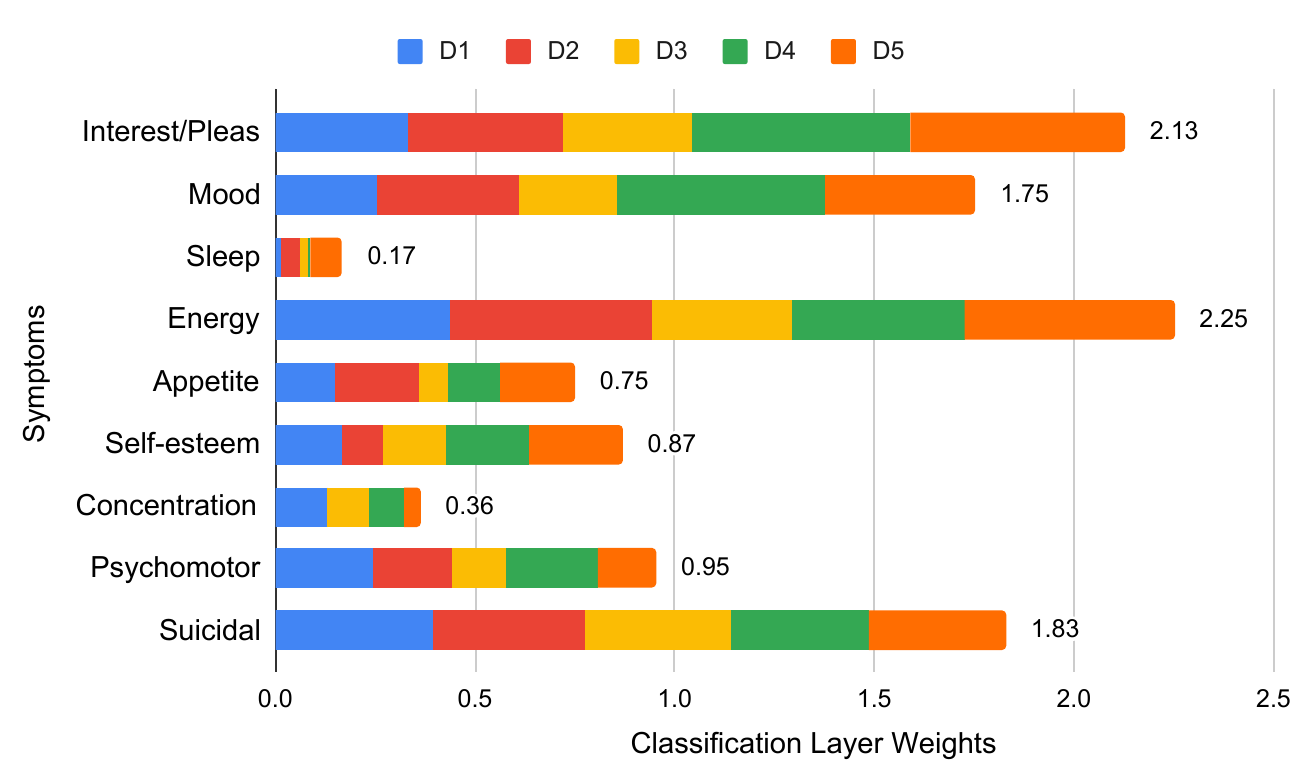}
    \caption{Classification weights (absolute values) for different symptoms over all datasets.}
    \label{fig:symptom_weights}
\end{figure}

\textbf{Result (3): Transparent Reasoning.}
Beyond accurate depression detection, ProtoDep offers valuable insights into its decision-making process through several avenues. Examining the weights assigned to various symptoms within its final layer unveils their relative importance in user classification. As illustrated in Figure \ref{fig:symptom_weights}, ProtoDep across diverse datasets prioritizes symptoms like "Fatigue or low energy" and "Lack of Interest," mirroring human expert judgment reported in \citep{yazdavar2017semi}. Interestingly, it assigns less weight to "Sleep Disorder" and "Concentration problems," potentially due to the ambiguity of these symptoms in textual data. For example, the tweet "lost in my own mind" might not explicitly mention keywords indicating "Concentration problems," making accurate classification challenging. This finding highlights the inherent difficulty in capturing nuanced depressive symptoms, even for human experts.

Furthermore, ProtoDep's user embedding layer with stacked attention layers holds promise for interpreting user classifications, similar to \citet{han2022hierarchical}. We analyzed attention scores to identify tweets that significantly influence user classification. However, echoing prior research \citep{bibal2022attention, wen2022revisiting, pruthi2020learning}, our extensive evaluation across both methods revealed no statistically significant association between attended tweets and those crucial for accurate classification. This suggests that while attention weights offer glimpses into model behavior, they might not directly explain specific classification outcomes in this context.

\section{Ablation Study}
\label{ablation}
We consider four different settings to validate the impact of different hyperparameters.

\textbf{Symptom Prototype Initialization.}
In this study, we explore alternative methods for initializing symptom prototypes. We compare two novel initialization approaches with the baseline method that utilizes a pre-trained language model (LLM) for symptom initialization. In the first setting, we leverage the ground truth lexicon as the foundation for symptom prototypes. We extract ground truth embeddings from this lexicon and then sample additional prototypes around them for each symptom class. A well-constructed lexicon captures domain-specific nuances and expert knowledge, which can enhance the quality of symptom prototypes. In the second setting, we depart from direct lexicon embeddings. Instead, we identify the nearest tweet in our dataset to each lexicon symptom and use its embedding as the basis for symptom prototypes. We then continue sampling around these tweet-grounded prototypes. By anchoring the initial prototypes to actual tweets, we aim to improve their relevance and alignment with real-world symptom expressions. Our experimental results in Table \ref{tab:ablation_init} demonstrate that both the lexicon-based and tweet-grounded initialization outperform the LLM baseline. Notably, the curated lexicon’s consideration of various combinations of the depression-indicative keywords contributes to its effectiveness. However, the marginal performance difference between these two settings and the baseline suggests that LLMs can achieve competitive results even in scenarios lacking human annotation or domain-specific knowledge.

\begin{table}[]
\begingroup
\setlength{\tabcolsep}{2pt} 
\renewcommand{\arraystretch}{1} 
\begin{tabular}{c|ccc}
\hline
\textbf{Loss Function} & \textbf{GPT-4} & \textbf{Lexicon} & \textbf{Lex. Tweets} \\ \hline
Triplet+MSE.+Ent.            & 0.990       & 0.990           & 0.989          \\
Sinkhorn+MSE.             & 0.947       & 0.969           & 0.954               \\
Sinkhorn             & 0.936       & 0.964           & 0.949                   \\ \hline
Avg. val. F1           & 0.958       & \textbf{0.975}  & 0.964                   \\ \hline
\end{tabular}
\endgroup
\caption{Comparison of average validation F1 scores for different initialization methods compared to different loss functions. "MSE." refers to Mean Squared Error loss, and "Ent." refers to Entropy loss.}
\label{tab:ablation_init}
\end{table}

\textbf{Prototype Loss Function.}
To assess the influence of different loss functions within the ProtoDep framework, we implemented two evaluation settings. The first setting, ProtoDep (Sinkhorn), exclusively employed the Sinkhorn loss to isolate the impact of the MSE loss in ProtoDep. The second setting, ProtoDep-Acc, combined Triplet loss with MSE and Entropy losses, mimicking conventional loss functions commonly used in prototype learning research.

As observed in Table \ref{tab:ablation_init}, ProtoDep-Acc achieved strong performance, demonstrating the capability of the ProtoDep framework. Notably, it surpassed other settings in terms of F1 scores. However, as illustrated in Table \ref{fig:pride}, while ProtoDep-Acc yielded higher F1 scores, its PRIDE scores indicated a failure to learn meaningful symptom-level prototypes. Conversely, ProtoDep (Sinkhorn) achieved better PRIDE scores, signifying successful prototype learning, but yielded lower F1 scores than ProtoDep.

This evaluation highlights a trade-off between classification performance and interpretability within the ProtoDep framework. While ProtoDep-Acc excelled in F1 scores, its learned prototypes lacked interpretability. In contrast, ProtoDep and ProtoDep (Sinkhorn) prioritized interpretability through meaningful prototypes but compromised classification accuracy. These findings suggest the need for careful consideration of loss function selection in balancing interpretability and performance within the ProtoDep framework.

\textbf{Attention Mechanism.}
Instead of employing the conventional user embedding attention mechanism, we introduce a single-layer Multi-head attention configuration. The outcomes are presented in Table \ref{tab:ablation_attention}.

\begin{table}[]
\begingroup
\setlength{\tabcolsep}{6pt} 
\renewcommand{\arraystretch}{1} 
\begin{tabular}{c|cc}
\hline
\textbf{Attention} & \textbf{Baseline} & \textbf{Multi-head}\\  \hline

Avg. val. F1        & \textbf{0.9407}      & 0.823    \\ \hline
\end{tabular}
\endgroup
\caption{Comparison of different attention mechanism.}
\label{tab:ablation_attention}
\end{table}

\begin{figure}
    \centering
    \includegraphics[width=1\linewidth]{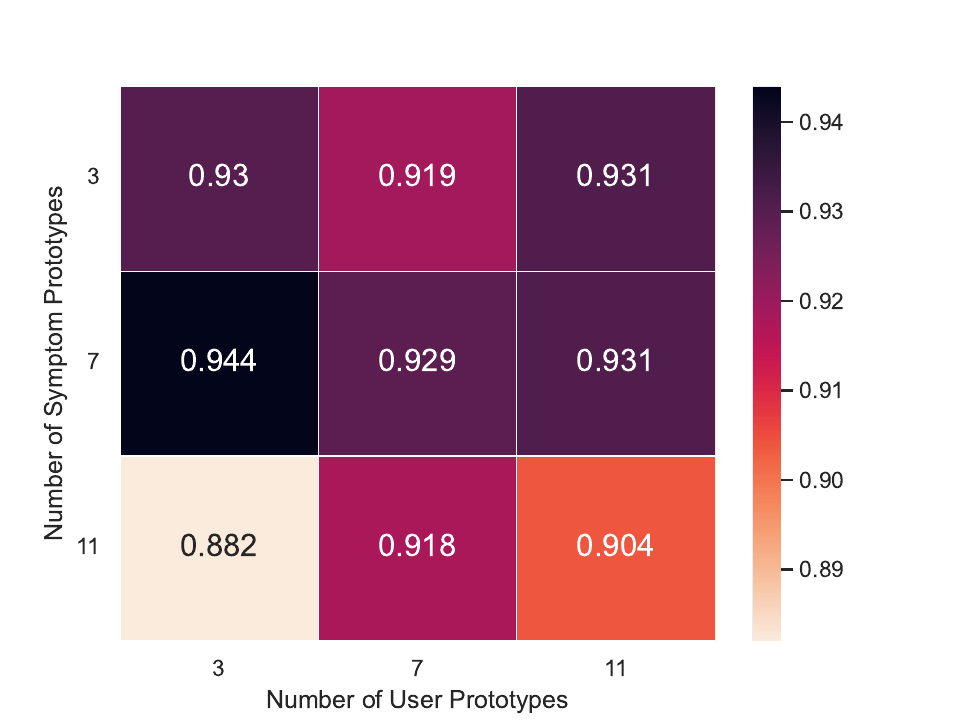}
    \caption{Average F1 score over all test datasets for different numbers of prototypes.}
    \label{fig:ablation_n_protos}
\end{figure}

\textbf{Number of Prototypes.}
Figure \ref{fig:ablation_n_protos} depicts the average F1 score across all five datasets, varying the number of prototypes. While the overall trend suggests that the model performs better with fewer prototypes, a nuanced examination reveals that individual datasets often favor a larger number of prototypes.

\section{Conclusion}

In this paper, we proposed ProtoDep, a novel framework that combines prototype learning and Large Language Models (LLMs) to provide explainable depression detection on Twitter. Unlike conventional "black-box" models, ProtoDep can generate transparent and interpretable explanations at three levels: symptom-level, case-based, and transparent decision-making weights. We evaluated ProtoDep on five benchmark datasets and showed it achieves competitive performance while learning meaningful and representative prototypes. We argue that ProtoDep has the potential to improve the trustworthiness and accountability of depression detection on social media, as well as to facilitate the understanding and intervention of mental health professionals. As a future work, we plan to investigate the applicability of ProtoDep to other social media platforms and mental health domains and enhance the explanation generation process with more clinical and contextual information. To sum up, ProtoDep is a novel and promising framework for explainable depression detection on social media, which can contribute to the well-being of individuals and society.

\section{Ethical Consideration}
We used a publicly available dataset introduced by \citet{han2022hierarchical}. Our investigation focuses exclusively on textual content, deliberately excluding user profile information. We emphasize the ethical implications of our model’s application and strongly discourage misuse compromising data security or privacy principles.

\section{Limitations}
While ProtoDep offers a promising approach to explainable depression detection on Twitter, it is essential to acknowledge potential limitations associated with this framework:

\textbf{Data Representativeness:} The performance and generalizability of ProtoDep heavily rely on the quality and representativeness of the training data. Biases or limitations within the training data can be reflected in the learned prototypes, potentially leading to biased predictions for demographics or groups underrepresented in the data.

\textbf{Tweet Encoder Model:} The effectiveness of the learned prototypes largely relies on the selection of the tweet encoder model. If the encoder model cannot accurately differentiate between various initial symptom prototypes, it will fail to converge to a meaningful prototype space at the end of the training process.

\textbf{Privacy Concerns:} Utilizing social media data for depression detection raises inherent privacy concerns. It is crucial to ensure user privacy throughout the data collection, processing, and explanation generation stages, adhering to relevant ethical and legal guidelines.

\textbf{Limited Scope:} While ProtoDep focuses on Twitter data, it might not generalize to other social media platforms with different content characteristics and user behaviors. Further research is needed to explore the framework's applicability across diverse platforms.

\textbf{Clinical Validation:} Although ProtoDep aims to aid mental health professionals by providing transparent case-based reasoning for each prediction, its effectiveness in real-world clinical settings requires rigorous validation through controlled studies with healthcare providers.

\textbf{Hyperparameter Tuning:} During our experiments with ProtoDep, we observed that model performance and prototype quality significantly varied for different hyperparameters. Careful hyperparameter optimization is necessary for adaptation to different domains. 

By acknowledging these limitations and actively working towards addressing them, future research can refine ProtoDep to ensure its responsible, ethical, and effective implementation in supporting mental health diagnosis and intervention.

\section*{Acknowledgements}
This research was funded by the National Science Foundation under Grant 2033521 A1, titled “KnowWhereGraph: Enriching and Linking Cross-Domain Knowledge Graphs using Spatially-Explicit AI Technologies.” The views, findings, conclusions, or recommendations presented in this paper are those of the authors and do not necessarily reflect the views of the National Science Foundation.





\end{document}